\documentclass[letterpaper, 10 pt, conference, twoside]{ieeeconf}
\makeatletter
\let\NAT@parse\undefined
\makeatother

\IEEEoverridecommandlockouts                              

\usepackage{moreverb,url}

\usepackage[colorlinks,bookmarksopen,bookmarksnumbered,citecolor=red,urlcolor=red]{hyperref}

\usepackage[numbers]{natbib}
\usepackage{multicol}
\usepackage{hyperref}
\usepackage[nolist,nohyperlinks]{acronym}
\usepackage{amsmath,amsfonts,amssymb}
\usepackage{tikz,tkz-euclide}
\usetikzlibrary{arrows,calc,shapes,positioning}
\tikzset{>=latex}
\usepackage{color,soul}
\usepackage{graphicx}
\usepackage{calc}
\usepackage{pgfplots}
\usepackage{colortbl}
\usepackage{todonotes}
\usepackage{multirow}
\usepackage{mathtools}
\def\identitymat{\mathbf{I}}

\newcommand{\units}[1]{\,\mathrm{#1}}

\def\ordinate{y}
\def\ordinatevec{\mathbf{y}}
\def\abscissa{x}
\def\abscissavec{\mathbf{x}}
\newcommand{\kernelraw}[1]{{{k}_{#1}}}
\newcommand{\kernel}[3]{{{k}_{#1}\left(#2,#3\right)}}
\newcommand{\kernelmat}[3]{{\mathbf{K}_{#1}(#2,#3)}}
\newcommand{\kernelvec}[3]{{\mathbf{k}_{#1}(#2,#3)}}

\newcommand\eventtime[1]{{t_{#1}}}  
\newcommand\eventtimevec{{\mathbf{t}}}  
\renewcommand\time[1]{{t_{#1}}}
\newcommand\frametime[1]{{\tau_{#1}}}

\def\costfunction{{C}}

\newcommand\trans[2]{{\mathbf{T}_{#1}^{#2}}}
\newcommand\homo[2]{{\mathbf{H}_{#1}^{#2}}}
\newcommand\proj[1]{{\pi_{#1}}}
\newcommand\rot[2]{{\mathbf{R}_{#1}^{#2}}}

\def\rotset{{\mathcal{R}}}

\newcommand\pos[2]{{\mathbf{p}_{#1}^{#2}}}
\def\posset{{\mathcal{P}}}
\newcommand\poscomponent[3]{{{}_{#3}{p}_{#1}^{#2}}}

\newcommand\statetime[1]{{\mathfrak{t}_{#1}}}
\def\nbstate{{Q}}

\newcommand\event[2]{{\mathbf{e}_{#1}^{#2}}}
\newcommand\eventset[2]{{\mathcal{E}_{#1}^{#2}}}
\newcommand\seevent[2]{{\bar{\mathbf{e}}_{#1}^{#2}}}
\def\coordinate{{\mathbf{e}}}
\def\eventfield{{f}}
\def\dfield{{d}}
\def\hfield{{g}}
\newcommand\hevent[2]{{\hat{\mathbf{e}}_{#1}^{#2}}}
\newcommand\eventcomponent[3]{{{}_{#3}{e}_{#1}^{#2}}}

\newcommand\rotang[2]{{r_{#1}^{#2}}}
\def\seed{\mathbf{s}}

\title{\LARGE \bf
Continuous-Time Gaussian Process Motion-Compensation for Event-vision Pattern Tracking with Distance Fields
}

\author{Cedric Le Gentil$^{1}$, Ignacio Alzugaray$^{2}$ and Teresa Vidal-Calleja$^{1}$
\thanks{This work  was partially supported  by  the  Australian  Research  Council  Discovery Project   under   Grant   DP210101336.}
\thanks{$^{1}$C. Le Gentil and T. Vidal-Calleja are with the Robotics Institute at the University of Technology Sydney, Australia: 
        {\tt\small \{cedric.legentil ; teresa.vidalcalleja\}@uts.edu.au}}%
\thanks{$^{2}$I. Alzugaray is with the Department of Computing, Imperial College London, UK: 
        {\tt\small i.alzugaray@imperial.ac.uk}}%
\thanks{\textcopyright 2023 IEEE. Personal use of this material is permitted. Permission from IEEE must be obtained for all other uses, in any current or future media, including reprinting/republishing this material for advertising or promotional purposes, creating new collective works, for resale or redistribution to servers or lists, or reuse of any copyrighted component of this work in other works
}
}

\begin{document}

\maketitle
\thispagestyle{empty}
\pagestyle{empty}

\begin{abstract}
This work addresses the issue of motion compensation and pattern tracking in event camera data.
An event camera generates asynchronous streams of \emph{events} triggered independently by each of the pixels upon changes in the observed intensity.
Providing great advantages in low-light and rapid-motion scenarios, such unconventional data present significant research challenges as traditional vision algorithms are not directly applicable to this sensing modality.
The proposed method decomposes the tracking problem into a local SE(2) motion-compensation step followed by a homography registration of small motion-compensated event batches.
The first component relies on Gaussian Process (GP) theory to model the continuous occupancy field of the events in the image plane and embed the camera trajectory in the covariance kernel function.
In doing so, estimating the trajectory is done similarly to GP hyperparameter learning by maximising the log marginal likelihood of the data.
The continuous occupancy fields are turned into distance fields and used as templates for homography-based registration.
By benchmarking the proposed method against other state-of-the-art techniques, we show that our open-source implementation performs high-accuracy motion compensation and produces high-quality tracks in real-world scenarios.

\end{abstract}

\section{Introduction}

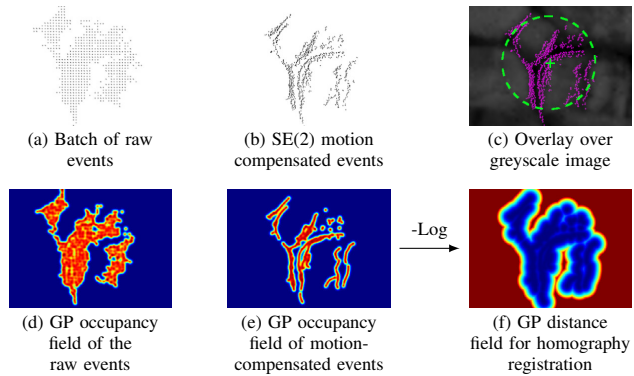
\begin{figure}
    \centering
    \def\scale{0.25}
    \def\legendspacing{-0.15cm}
    \def\vertspacing{0.6cm}
    \def\horispacing{0.5cm}
    \def\lasthorispacing{0.8cm}
    \begin{tikzpicture}
        \tikzstyle{legend} = [rectangle, text width=\scale\columnwidth, align=center, anchor=north, font=\scriptsize]
        \node (raw) {\includegraphics[clip, trim=0cm 0.25cm 0cm 0.5cm, width=\scale\columnwidth]{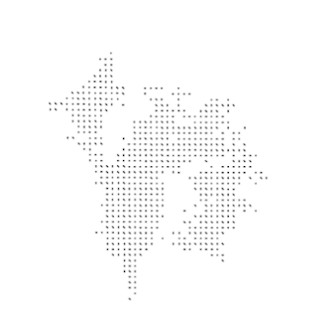}};
        \node[below=\vertspacing of raw] (rawfield) {\includegraphics[clip, trim=0cm 0.25cm 0cm 0.5cm, width=\scale\columnwidth]{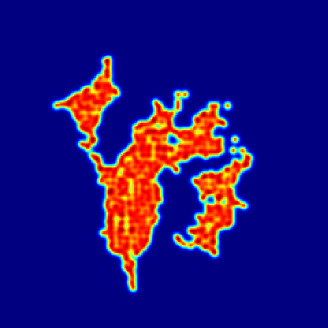}};
        \node[right=\horispacing of raw] (undist) {\includegraphics[clip, trim=0cm 0.25cm 0cm 0.5cm, width=\scale\columnwidth]{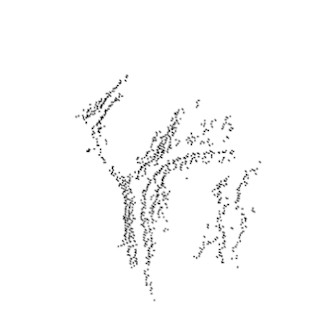}};
        \node[right=\horispacing of rawfield] (undistfield) {\includegraphics[clip, trim=0cm 0.25cm 0cm 0.5cm, width=\scale\columnwidth]{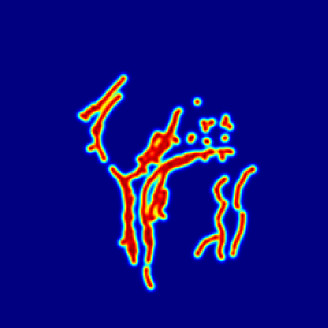}};
        \node[right=\lasthorispacing of undist] (greyscale) {\includegraphics[clip, trim=0cm 0.25cm 0cm 0.5cm, width=\scale\columnwidth]{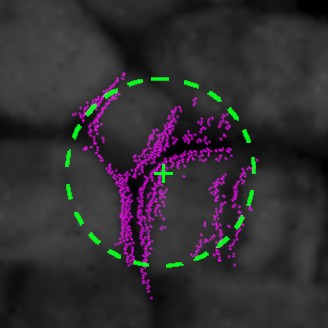}};
        \node[right=\lasthorispacing of undistfield] (distfield) {\includegraphics[clip, trim=0cm 0.25cm 0cm 0.5cm, width=\scale\columnwidth]{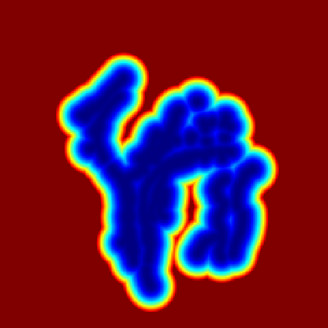}};

        \node[legend, below=\legendspacing of raw]        {(a) Batch of raw\\[0pt]events};
        \node[legend, below=\legendspacing of undist]     {(b) SE(2) motion\\[0pt]compensated events};
        \node[legend, below=\legendspacing of greyscale]  {(c) Overlay over\\[0pt]greyscale image};
        \node[legend, below=\legendspacing of rawfield]   {(d) GP occupancy\\[0pt]field of the\\[0pt]raw events};
        \node[legend, below=\legendspacing of undistfield]{(e) GP occupancy\\[0pt]field of motion-\\[0pt]compensated events};
        \node[legend, below=\legendspacing of distfield]  {(f) GP distance\\[0pt]field for homography\\[0pt]registration};

        \draw[->] (undistfield) -- node[above] {\scriptsize -Log} (distfield);
    \end{tikzpicture}
    \vspace{-0.35cm}

    \caption{The proposed pattern tracking method performs motion-compensation of event batches ((a) and (b)) using a Gaussian Process (GP) continuous representation of the events' occurrences in the image plane (occupancy fields in (d) and (e)). Applying the logarithm to the occupancy field yields a continuous distance field (f) later used for homography registration of different event batches. For visualisation only: (e) shows the compensated events over the corresponding greyscale image (green circle is the area of interest), and the colours in (f) are saturated above a certain threshold to show the details in the vicinity of the events.}
    \label{fig:fields}
\end{figure}

In the context of \ac{slam}, robotic systems have been built upon a variety of exteroceptive sensors throughout the years.
Vision-based \ac{slam} systems generally consist of two main components that are the front-end, dealing with visual keypoints detection, tracking and/or matching; and the back-end, which estimates the system's pose and keypoints location in 3D.
Recently, the robotics community started to investigate the use of \emph{event cameras} for state estimation.
Unlike standard ``frame-based" cameras that trigger data acquisition of millions of pixels at specific, discrete timestamps, event cameras do not provide regular snapshots of the environment.
They produce an asynchronous stream of \emph{events} driven by changes of intensity observed independently by each of the cameras' pixels.
Each event corresponds to a tuple of four values that represents the direction of the light intensity change compared to a given threshold (referred to as polarity), the timestamp of that change, and the $x$-$y$ location of the pixel in the image plane.
Note that, as event cameras only react to changes, events generated in a static environment with constant illumination are produced solely due to the motion of the camera and concentrate on highly salient scene regions projected on the image plane.
Reacting only to small and incremental changes, event cameras exhibit an extended dynamic range and extreme resilience to motion blur in comparison with their frame-based counterparts.
This unconventional sensing modality makes event-based state estimation an open research topic where mature techniques for frame-based vision cannot be directly applied, especially when it comes to the front-end side of \ac{slam}.
In this paper, we propose a novel method to address the issue of pattern tracking in the event stream.

As event generation and camera motion are correlated, the proposed method models the events-only pattern tracking as a motion estimation problem.
This is also commonly denoted as motion compensation, or unwarping.
Our method first corrects locally the flow of events in small batches assuming \ac{SE2} motion in the image plane.
By modelling the spatiotemporal ($x$-$y$-time) occupancy of the events with a \ac{gp}  that embeds the camera motion in its covariance kernel, the method estimates the \ac{SE2} trajectory by maximising the log marginal likelihood of the data similarly to the standard kernel hyperparameter learning.
Fig.~\ref{fig:fields} shows the event occupancy field at given timestamps before and after motion estimation.
Turning an occupancy field into a distance field (Fig.~\ref{fig:fields}(f)) allows for the query of the distance to the closest motion-compensated events at any location in the image plane.
Long-term pattern tracking is  achieved by estimating homography between batches of \ac{SE2}-motion-compensated events employing the above-mentioned distance fields.
The contributions associated with this work are:
\begin{itemize}
    \item The formulation of the continuous-time motion-compensation of event data as a hyperparameter learning optimisation based on the \ac{gp} theory.
    \item The introduction of a patch registration algorithm for event data using continuous \ac{gp} distance fields.
    \item A set of experiments analysing the performance of the proposed, open-source method \footnote{https://github.com/UTS-CAS/gp\_comet} compared to other open-source state-of-the-art alternatives.
\end{itemize}

\section{Related Work}
Monocular event-based \ac{slam} and \ac{vo} have been active topics of research since the conception of event cameras. 
While early approaches on the topic tackle simplified scene geometry and camera motions \cite{weikersdorfer2012, kim2014}, first works achieving 6-\ac{dof} event-based monocular \ac{slam} and \ac{vo} \cite{kim2016, rebecq2017a} opt for a direct approach where the definition of explicit visual features is avoided and thus the need for data association is circumvented.
Event-based visual features are arguably more challenging to be detected and tracked than their traditional frame-based counterpart as motion and appearance are intertwined in the event stream \cite{alzugaray2022}.
Consequentially, many approaches to indirect, feature-based \ac{vo} for event cameras rely on complementary sensing modalities such as traditional frame-based cameras in \cite{tedaldi2016, kueng2016, li2019a}, where corners and edges are extracted from intensity images and used as templates to be tracked in the event stream. 
More complex approaches such as \cite{gehrig2018, gehrig2020} employ generative models \cite{gallego2015} that correlate motion, intensity and event data generation, leading to the definition of feature-based \ac{vo} systems such as the one described in \cite{hidalgo-carrio2022}.

Several works have devised the use of image-like event-frames or other time-invariant representations \cite{jiao2021} on which traditional frame-based techniques to feature detection and tracking can be easily adapted. 
For instance, the method from \cite{chiberre2021} infers intensity gradient images from batches of events whereas the algorithm in \cite{glover2021} generates image-like edge-maps that are updated at each new event and on which traditional Harris corners \cite{harris1988} are detected.
Similarly, the event-based \ac{vo} systems described in \cite{rebecq2017,rosinolvidal2018} use an \ac{imu}-aided motion-compensation scheme to also generate event-frames where traditional frame-based FAST corners \cite{rosten2006} are detected and tracked using KLT \cite{lucas1981}.
Initially explored in \cite{gallego2017} and further formalised in \cite{stoffregen2019, gallego2019}, motion-compensation is one of the most prevalent frameworks in the event-based literature as it correlates motion and event generation.
Similarly to \cite{rebecq2017}, the \ac{vo} approach presented in  \cite{zhu2017} also relies on event-frames to detect \cite{harris1988} corners, but establishes data association based on a per-feature \ac{em} motion-compensation scheme \cite{zhu2017b}.
Approaches such as \cite{seok2020, chui2021} also apply motion-compensation techniques to track features in the event stream, but model the feature's motion using a continuous-time formulation in contrast to \cite{zhu2017, zhu2017b, rebecq2017}.
As event motion-compensation frameworks increase in popularity, other similar alternatives have been proposed emphasizing specific aspects such as robustness \cite{nunes2020, xu2020}, efficiency \cite{nunes2021} or global optimality \cite{liu2020,peng2020}.

Significant research effort has been also dedicated to the definition of event-driven features \cite{alzugaray2022}, i.e. features that can be detected and tracked by processing each event individually and without any form of event batching.
While most of the literature on this topic is focused on corner-events, their detection \cite{vasco2016,mueggler2017,alzugaray2018,li2019,scheerlinck2019, manderscheid2019, yilmaz2021} and tracking \cite{clady2015,clady2017, alzugaray2018a, li2021,dai2022tightly}, corner-events are still to be successfully applied to \ac{vo} due to their limited reliability.
Nonetheless, emerging event-driven pattern-tracking techniques as the ones described in \cite{alzugaray2019,alzugaray2020} have found their way to \ac{vo} systems as the ones described in \cite{mahlknecht2022, liu2022}.
Another promising approach is the use of event-based line features \cite{brandli2016, dardelet2018, peng2021continuous, tschopp2021} which exploits the natural sensitivity of event cameras to edges and for which a significant number of event-based \ac{vo} systems have been proposed in recent years \cite{legentil2020, chamorro2020, chamorro2022}.

Among all these event-vision works, the proposed method finds similarities in the continuous-time motion-compensation of \cite{seok2020}, and the kernel-based optimisation in \cite{nunes2021} where the kernel represents a ``potential" and the entropy of the data is minimised.
Our method leverages \acp{gp} regression \cite{Rasmussen2006} over \ac{SE2} to characterise the continuous motion as opposed to a 2-\acp{dof} (translation in the image plane) Bezier parameterisation in \cite{seok2020}.
Various works use \acp{gp} to perform efficient long-term state estimation using motion priors to derive exactly sparse \ac{gp} kernels as in \cite{Barfoot2014,Anderson2015,liu2022}.
In the proposed method, we opted for an approach similar to \cite{LeGentil2021b} where the inducing values of the \ac{gp} state are estimated.
Embedding these \ac{gp} trajectories in a spatiotemporal kernel that characterises the covariance between events, we can optimise the image-plane motion by maximising the marginal likelihood of the events as for the traditional \ac{gp} regression hyperparameter learning \cite{Rasmussen2006}.
When employing the proposed motion-compensated scheme to pattern tracking, a key element of the process is the proposed homography registration algorithm that relies on continuous \ac{gp}-based distance fields.
Given the \ac{gp} event occupancy field in the \ac{SE2}-corrected patches, distance fields can be obtained by simply applying a negative logarithm as shown in \cite{Wu2021} in the context of implicit surface \cite{Williams2006, Lee2019a}.

\section{Method}
The proposed method addresses the motion compensation of events and, specifically, applies it to the problem of pattern tracking.
Starting from a given seed ($x$-$y$ coordinates in the image plane at a given time $t$), our pipeline first performs motion compensation using a \ac{SE2} motion model over a small batch of events in the spatio-temporal vicinity of the seed.
The motion-compensated events are used to create a template that will be used for later batch-to-template registration.
The seed location is updated based on the estimated motion and is used to collect the next batch of events.
For the second batch and the subsequent ones, the motion compensation is performed followed by a homography registration with respect to the previous batch and the template.
The template is dynamically updated with the last registered motion-compensated event batch.
The data pipeline of the proposed method is shown in Fig.~\ref{fig:method_diagram}.
This framework relies on the assumption that a continuous-time \ac{SE2} motion in the image plane can explain the generation of events over small periods of time (i.e., locally), and that homography transformations are sufficient to model the appearance changes of patches over longer time periods (i.e., globally).

\subsection{GP regression preliminaries}
\ac{gp} regression \cite{Rasmussen2006} is a kernel-based, probabilistic interpolation method that allows for data-driven inferences of an unknown signal given a set of noisy observations.
The so-called kernel function represents the covariance between two instantiations of the signal and is essentially the only assumption about the signal.
Let us consider $h(\abscissa)$ as a function of $\abscissa \in \mathbb{R}$, with noisy observations $\ordinate_i = h(\abscissa_i) + \eta$ where $\eta \sim \mathcal{N}(0,\sigma^2_{\ordinate})$ and $i = 1,\cdots,N$.
By modelling $h(\abscissa)$ with a \ac{gp} as $h(\abscissa) \sim \mathcal{GP}\big(0,\kernel{h}{\abscissa}{\abscissa'} \big)$, with $\kernel{h}{\abscissa}{\abscissa'} = \text{cov}\big(h(\abscissa),h(\abscissa')\big)$ being the covariance kernel, it is possible to infer any new point $ h^*(\abscissa)$ and its variance $\text{var}\big(h^*(\abscissa)\big)$ as
\begin{align}
    h^*(\abscissa)& \sim \mathcal{N}\big(\kernelvec{h}{\abscissa}{\abscissavec}\big[\kernelmat{h}{\abscissavec}{\abscissavec} + \sigma_y^2\identitymat\big]^{-1}\ordinatevec,
    \label{eq:standard_gp_regression}
    \\
    &\kernel{h}{\abscissa}{\abscissa}\text{-}\ \kernelvec{h}{\abscissa}{\abscissavec}\big[\kernelmat{h}{\abscissavec}{\abscissavec} + \sigma_y^2\identitymat\big]^{-1}\kernelvec{h}{\abscissavec}{\abscissa}\big),
    \nonumber
\end{align}
where $\ordinatevec = \begin{bmatrix} \ordinate_1& \cdots & \ordinate_N) \end{bmatrix}^\top$, $\abscissavec = \begin{bmatrix} \abscissa_1 & \cdots & \abscissa_N \end{bmatrix}^\top$, $\kernelvec{h}{\abscissa}{\abscissavec} = \begin{bmatrix}   \kernel{h}{\abscissa}{\abscissa_1} & \cdots & \kernel{h}{\abscissa}{\abscissa_N} \end{bmatrix}$, $\kernelvec{h}{\abscissavec}{\abscissa} = \kernelvec{h}{\abscissa}{\abscissavec}^\top$, and $\kernelmat{h}{\abscissavec}{\abscissavec} = \begin{bmatrix} \kernelvec{h}{\abscissavec}{\abscissa_1} & \cdots & \kernelvec{h}{\abscissavec}{\abscissa_N} \end{bmatrix}$.
Note that in the rest of this paper, the vector $\ordinatevec$ will be also referred to as the set of \emph{inducing values}.

The kernel function generally depends on a set of \emph{hyperparameters} that we denote $\Theta_{h}$.
For example, the square exponential kernel possesses two hyperparameters which are a scaling factor and a lengthscale.
While the hyperparameters can be set arbitrarily with an educated guess, they can also be learnt directly from the data.
An approach to hyperparameter learning is to maximise the log marginal likelihood of the observations of $h$ with respect to the value of the hyperparameters, $\underset{\Theta_{h}}{\text{argmax}}\ \log(P(\abscissavec \lvert \ordinatevec, \Theta_{h}) )$, with
\begin{align}
\begin{aligned}
    \log(P(\ordinatevec \lvert \abscissavec, \Theta_{h}) ) =& -\frac{1}{2} \ordinatevec^\top (\kernelmat{h}{\abscissavec}{\abscissavec}+\sigma_y^2\identitymat)^{-1}\ordinatevec
    \\
    - \frac{1}{2}\log(\lvert& \kernelmat{h}{\abscissavec}{\abscissavec}+\sigma_y^2\identitymat \lvert) 
    - \frac{N}{2}\log(2\pi).
    \label{eq:hyperparam_learning}
\end{aligned}
\end{align}

For more details about \ac{gp} regression and hyperparameter learning, we invite the reader to refer to \cite{Rasmussen2006}.

\subsection{System's description and definitions}

Let us consider an event camera providing events at timestamps $\eventtime{i}$ ($i = 1,\dots, N$).
For each event, we denote $\event{\eventtime{i}}{i}$ its $x$-$y$ coordinate vector in the image plane at time $\eventtime{i}$.
The subscript denotes the timestamp in which the coordinates of the event are expressed.
Locally assuming a \ac{SE2} motion in the image plane, and given some knowledge of that motion, an event can be transformed to a different timestamp as $[\seevent{\eventtime{j}}{i}{}^\top$ $1]^\top$ $=$ $\trans{\eventtime{j}}{\eventtime{i}}[\event{\eventtime{i}}{i}{}^\top$ $1]^\top$,
with $\seevent{\eventtime{j}}{i}$ the \ac{SE2}-transformed event, and $\trans{\eventtime{j}}{\eventtime{i}} \in \mathbb{R}^{3\times3}$ the homogeneous transformation matrix that characterises the \ac{SE2} motion in the image plane between timestamps $\eventtime{j}$ and $\eventtime{i}$.
This transformation matrix is associated with the rotation $\rotang{\eventtime{j}}{\eventtime{i}}$ and translation $\pos{\eventtime{j}}{\eventtime{i}}$ $=$ $[\poscomponent{\eventtime{j}}{\eventtime{i}}{x}$ $\poscomponent{\eventtime{j}}{\eventtime{i}}{y}]^\top$ as 
$\trans{\eventtime{j}}{\eventtime{i}}$ $=$ $\begin{bsmallmatrix} \rot{\eventtime{j}}{\eventtime{i}} & \pos{\eventtime{j}}{\eventtime{i}} \\ \mathbf{0}  & 1\end{bsmallmatrix}, \text{with}\quad \rot{\eventtime{j}}{\eventtime{i}} = \begin{bsmallmatrix} \cos(\rotang{\eventtime{j}}{\eventtime{i}}) & -\sin(\rotang{\eventtime{j}}{\eventtime{i}}) \\ \sin(\rotang{\eventtime{j}}{\eventtime{i}}) &  \cos(\rotang{\eventtime{j}}{\eventtime{i}})\end{bsmallmatrix}$.

Similarly, an event $\event{\eventtime{i}}{i}$ can be projected via homography on the image plane, $\hevent{\eventtime{j}}{i}$, at any timestamp $\eventtime{j}$ as $\hevent{\eventtime{j}}{i}$~$=$~$\proj{\homo{\eventtime{j}}{\eventtime{i}}}(\event{\eventtime{i}}{i})$~$=$~$[\frac{\eventcomponent{\eventtime{j}}{i}{x}}{\eventcomponent{\eventtime{j}}{i}{w}}$~$\frac{\eventcomponent{\eventtime{j}}{i}{y}}{\eventcomponent{\eventtime{j}}{i}{w}}]^\top$, with $[\eventcomponent{\eventtime{j}}{i}{x}$~$\eventcomponent{\eventtime{j}}{i}{y}$~$\eventcomponent{\eventtime{j}}{i}{w}]^\top$~$=$~$\homo{\eventtime{j}}{\eventtime{i}}[\event{\eventtime{i}}{i}$~$1]^\top$,
and $\homo{\eventtime{j}}{\eventtime{i}} \in \mathbb{R}^{3\times3}$ the homograhy matrix.

\subsection{Continuous-time SE(2) event data motion-compensation}

\begin{figure}
    \centering
    \def\scale{0.5}
    \def\topspacing{-0.2cm}
    \def\bottomspacing{-0.1cm}
    \def\horispacing{0.90cm}
    \def\vertspacing{0.9cm}
    \def\smallvertspacing{0.25cm}
    \begin{tikzpicture}
        \tikzstyle{block} = [draw, fill=white, rectangle, minimum height = 2em, text width = 7.5em,  minimum width = 7.5em, align = center, node distance = 14em]
        \tikzstyle{input} = [inner sep=0,outer sep=0, fill=white, rectangle, minimum height = 0.1em, text width = 7.5em,  minimum width = 7.5em, align = center, node distance = 5em, draw=none]
        \tikzstyle{blocklarge} = [draw, fill=white, rectangle, minimum height = 1.5em, text width = 22.5em,  minimum width = 22.5em, align = center, node distance = 8.5em]
        \tikzstyle{decision} = [draw, fill=white, aspect=2, diamond, minimum height = 3em, text width = 5.2em,  minimum width = 3em, align = center, node distance = 10em]
        \tikzstyle{output} = [inner sep=0,outer sep=0, fill=white, minimum height = 0.5em, text width = 20.0em,  minimum width = 20.0em, align = center, node distance = 11em, draw=none]
        \def\linedistoffset{8em}

        \node [input] (rawev) {\scriptsize Raw event batches $\eventset{\frametime{n}}{}$};
        \node [block, below= \smallvertspacing of rawev] (se) {\scriptsize Local \textbf{SE(2) motion\\[0pt]compensation} (continuous\\[0pt]GP trajectory in\\[-4pt]image space) };
        \node [blocklarge, below=\vertspacing of se, xshift=7.5em] (hom) {\scriptsize \textbf{Homography registration} with respect to $\bar{\eventset{}{}}_{\frametime{n-1}}$\\[-4pt]and $\hat{\eventset{}{}}_{T}$ (GP continuous distance fields)};
        \node [block, above=\smallvertspacing of hom, xshift=7.5em] (dtemp) {\scriptsize \textbf{Dynamic template}\\[-4pt]computation $\hat{\eventset{}{}}_{T}$};
        \node [output, below=\smallvertspacing of hom] (res) {\scriptsize  Track position update and registered events $\hat{\eventset{}{}}_{\frametime{n}}$ };

        \draw[->] (rawev) -- (se);
        \draw[->] (se.south) -- node[left]{\scriptsize $\bar{\eventset{}{}}_{\frametime{n}}$} (se.south |- hom.north);
        \draw[->] (hom) -- (res);
        \coordinate[right=0.2cm of hom] (j1);
        \coordinate[right=0.2cm of dtemp] (j2);
        \draw[->] (hom.east) -- (j1) -- node[right]{\scriptsize $\hat{\eventset{}{}}_{\frametime{n}}$} (j2) -- (dtemp.east);
        \draw[->] (dtemp.west) -| node[above]{\scriptsize $\hat{\eventset{}{}}_{T}$} (hom.north);

        \node[right=\horispacing of rawev, yshift=-0.4cm] (graph) {\includegraphics[clip, trim=0cm 0cm 0cm 0cm, width=\scale\columnwidth]{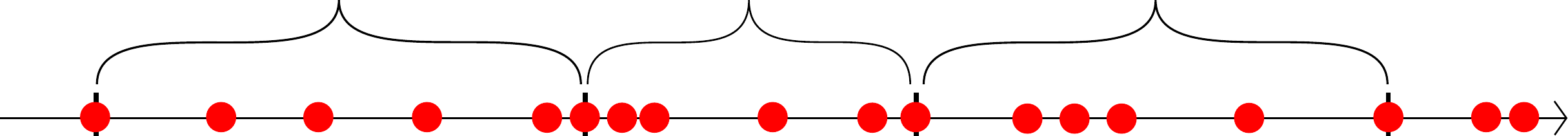}};
        \node[anchor=south, above=\topspacing of graph, xshift=\scale*-2.5cm] {\small $\eventset{\frametime{n-1}}{}$};
        \node[anchor=south, above=\topspacing of graph, xshift=\scale*-0.2cm] {\small $\eventset{\frametime{n}}{}$};
        \node[anchor=south, above=\topspacing of graph, xshift=\scale*2.0cm] {\small $\eventset{\frametime{n+1}}{}$};
        \node[anchor=north, below=\bottomspacing of graph, xshift=\scale*-3.8cm] {\small $\frametime{n-1}$};
        \node[anchor=north, below=\bottomspacing of graph, xshift=\scale*-1.1cm] {\small $\frametime{n}$};
        \node[anchor=north, below=\bottomspacing of graph, xshift=\scale*0.7cm] {\small $\frametime{n+1}$};
        \node[anchor=north, below=\bottomspacing of graph, xshift=\scale*3.2cm] {\small $\frametime{n+2}$};
        \node[anchor=north, below=\bottomspacing of graph, yshift=0.5cm, xshift=\scale*4.4cm] {\small $\eventtime{}$};
        \node[anchor=north, below=\bottomspacing of graph, yshift=-0.3cm] {\small Timeline of fixed-size event batches};
    \end{tikzpicture}
    \vspace{-0.35cm}
    \caption{Overview of the proposed event-only pattern tracking method (event timeline in upper right with red dots as events' timestamps).}
    \label{fig:method_diagram}
\end{figure}
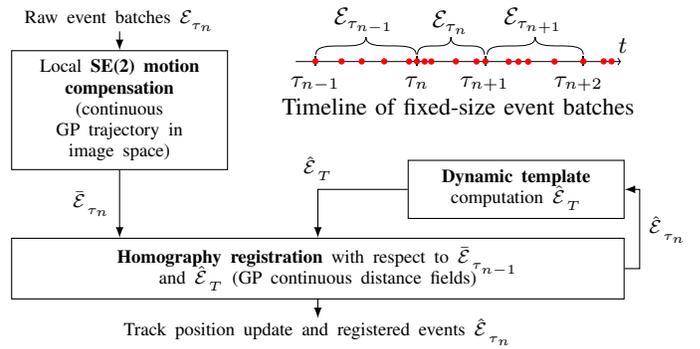

This subsection details the proposed \ac{SE2} motion-compensation mechanism for a set of raw events $\eventset{\frametime{n}}{}$.
This set contains a fixed number of events occurring after time $\frametime{n}$ as illustrated in Fig.~\ref{fig:method_diagram}.
Based on the assumption that the events in $\eventset{\frametime{n}}{}$ are generated by a \ac{SE2} camera motion parallel to the image plane, let us model the trajectory rotation $\rotang{\frametime{n}}{\time{}}$ and position $\pos{\frametime{n}}{\time{}}$ with three independent \acp{gp}
\begin{align}
    \rotang{\frametime{n}}{\time{}}(\time{})& \sim \mathcal{GP}\left(0, \kernel{\rotang{}{}}{\time{}}{\time{}'}\right)
    \label{eq:rot}
    \\
    \poscomponent{\frametime{n}}{\time{}}{\bullet}(\time{})& \sim \mathcal{GP}\left(0, \kernel{\poscomponent{}{}{\bullet}}{\time{}}{\time{}'}\right),
    \label{eq:pos}
\end{align}
where $\kernelraw{r}$ and $\kernelraw{p}$ are the associated covariance kernels.
The \emph{inducing values} of these \acp{gp} (i.e. the set of ``observations" $\ordinate_i$ in \eqref{eq:standard_gp_regression} at arbitrarily chosen timestamps $\statetime{i}$ with $i=1,\cdots,\nbstate$) are denoted $\rotset_n$ and $\posset_n$.
The associated continuous-time transformation matrix is denoted $\trans{\frametime{n}}{\time{}}(\time{})$, which is ``parameterised" by $\rotset_n$ and $\posset_n$.

Now, let us model the occurrence of events with another \ac{gp} (in an implicit-surface-like manner) with 
\begin{align}
    \eventfield\left(\begin{bmatrix}\coordinate \\ \time{}\end{bmatrix}\right) \sim \mathcal{GP}\left(0, \kernel{\eventfield}{\begin{bmatrix}\coordinate \\ \time{}\end{bmatrix}}{\begin{bmatrix}\coordinate' \\ \time{}'\end{bmatrix}}\right).
    \label{eq:gp_se_field}
\end{align}
The associated observations are the event coordinates and timestamps in $\eventset{\frametime{n}}{}$ as inputs and the constant value \emph{one} as outputs.
Thus, $\eventfield$ represents the continuous occupancy field of the image plane after motion-compensation (according to a continuous-time \ac{SE2} model) of the event stream.
This is illustrated in Fig.~\ref{fig:fields}~(a,b,d,e).
The kernel function $\kernelraw{\eventfield}$ embeds the continuous-time \ac{SE2} trajectory defined in \eqref{eq:rot} and \eqref{eq:pos}:
\begin{align}
    \kernel{\eventfield}{\begin{bmatrix}\coordinate \\ \time{}\end{bmatrix}}{\begin{bmatrix}\coordinate' \\ \time{}'\end{bmatrix}} = \sigma_{\eventfield{}} \exp\left(-\frac{\lVert\trans{\frametime{n}}{\time{}}(\time{})\coordinate - \trans{\frametime{n}}{\time{}'}(\time{}')\coordinate' \rVert^2}{2l_{\eventfield{}}^2} \right).
    \nonumber
\end{align}
Intuitively the kernel $\kernelraw{\eventfield}$ corresponds to a standard square exponential kernel over motion-compensated events.
As this kernel relies on the knowledge of the \ac{SE2} trajectory, it depends on the \ac{gp} inducing values in $\rotset_n$ and $\posset_n$.
In other words, the inducing values in $\rotset_n$ and $\posset_n$ can be considered as hyperparameters of the kernel function $\kernelraw{\eventfield}$.
The proposed \ac{SE2} compensation scheme optimises the trajectory as a hyperparameter-learning problem by maximising the marginal likelihood of the event data given the hyperparameters $\Theta_{n} = \{\rotset_n, \posset_n\}$ as in \eqref{eq:hyperparam_learning}:
\begin{align}
    \Theta_{n}^* = \underset{\Theta_{n}}{\text{argmax}}\ \log(P(\ordinatevec \lvert \eventset{}{}, \eventtimevec, \Theta_{n}) ),
    \label{eq:likelihood_max}
\end{align}
with $\eventtimevec$ the vector of event timestamps in $\eventset{\frametime{n}}{}$.
In the rest of this paper, $\bar{\eventset{}{}}_{\frametime{n}}$ will refer to the set of events in ${\eventset{}{}}_{\frametime{n}}$ after \ac{SE2} motion-compensation using the estimated trajectory parameters $\Theta_{n}^*$, \eqref{eq:rot}, and \eqref{eq:pos}.

\subsection{Homography registration}

Let us consider two sets of locally-\ac{SE2}-compensated events $\bar{\eventset{}{}}_{\frametime{A}}$ and $\bar{\eventset{}{}}_{\frametime{B}}$.
Assuming that both sets contain observations of a single planar patch, the proposed method estimates the homography transformation $\homo{\frametime{A}}{\frametime{B}}$ between the two sets using distance fields in the image space.
Similarly to \eqref{eq:gp_se_field}, we model the continuous occupancy field associated to $\bar{\eventset{}{}}_{\frametime{A}}$ as $\hfield(\coordinate) \sim \mathcal{GP}(0, \kernel{\hfield}{\coordinate}{\coordinate'})$ with $\kernelraw{\hfield}$ the corresponding spatial covariance kernel.
It is possible to infer a continuous distance field $\dfield$ as the negative logarithm of the occupancy field $\hfield$ similarly to \cite{Wu2021} or \cite{legentil2023accurate}: $\dfield(\coordinate) = -\log(\hfield(\coordinate))$.
Note that the occupancy field $\hfield(\coordinate)$ is providing values between zero (\ac{gp} means) and one (observation values).
As illustrated in Fig.~\ref{fig:fields}~(e,f), taking the negative logarithm of such a field provides a distance-to-closest-event field that is equal to zero around events and growing monotonously when inferring further.
By querying the distance field for the events in $\bar{\eventset{}{}}_{\frametime{B}}$ projected at time $\frametime{A}$ with $\homo{\frametime{A}}{\frametime{B}}$, it is possible to build a non-linear least-square cost function $\costfunction$ to register $\bar{\eventset{}{}}_{\frametime{B}}$ to $\bar{\eventset{}{}}_{\frametime{A}}$ by estimating $\homo{\frametime{A}}{\frametime{B}}$:
\begin{align}
    \costfunction(\homo{\frametime{A}}{\frametime{B}}) = \sum_{\seevent{\frametime{B}}{i} \in \bar{\eventset{}{}}_{\frametime{B}}} (\dfield( \proj{\homo{\frametime{A}}{\frametime{B}}}(\seevent{\frametime{B}}{i}))^2).
    \label{eq:h_cost}
\end{align}

\subsection{Pattern Tracking}

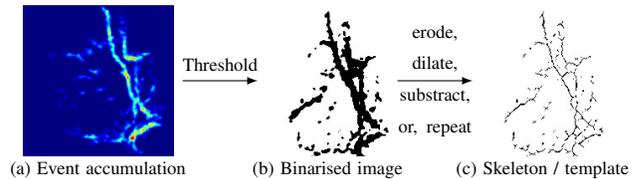
\begin{figure}
    \centering
    \def\scale{0.30}
    \def\legendspacing{-0.05cm}
    \def\vertspacing{0.6cm}
    \def\horispacing{1.0cm}
    \def\lasthorispacing{1.0cm}
    \def\imgheight{2cm}
    \begin{tikzpicture}
        \tikzstyle{legend} = [inner sep=0, outer sep=0, rectangle, text width=\scale\columnwidth, align=center, anchor=north, font=\scriptsize]
        \node (raw) {\includegraphics[clip, trim=0cm 0.5cm 0.5cm 0.cm, height=\imgheight]{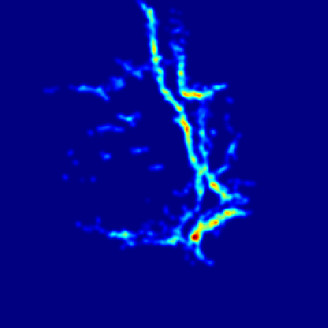}};
        \node[right=\horispacing of raw] (bin) {\includegraphics[clip, trim=0.3cm 0.5cm 0.5cm 0.0cm, height=\imgheight]{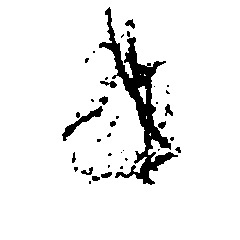}};
        \node[right=\lasthorispacing of bin] (skel) {\includegraphics[clip, trim=0.3cm 0.5cm 0.5cm 0.0cm, height=\imgheight]{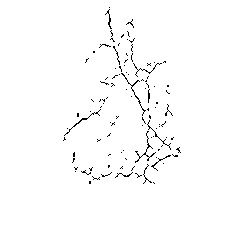}};

        \node[legend, below=\legendspacing of raw]        {(a) Event accumulation};
        \node[legend, below=\legendspacing of bin]     {(b) Binarised image};
        \node[legend, below=\legendspacing of skel]  {(c) Skeleton / template};

        \draw[->] (raw) -- node[above] {\scriptsize Threshold} (bin);
        \draw[->] (bin) -- node[text width=1cm, align=center ]{\scriptsize erode, dilate, substract, or, repeat} (skel);
    \end{tikzpicture}
    \vspace{-0.2cm}

    \caption{Creation process of the dynamic template using skeletonisation.}
    \label{fig:dynamic_template}
\end{figure}

As illustrated in Fig.~\ref{fig:method_diagram}, the proposed method combines the \ac{SE2} continuous-time motion-compensation to locally compensate event batches and then performs discrete homography registration between batches to track patterns over time.
Performing solely sequential \emph{batch-to-batch} homography registration between $\bar{\eventset{}{}}_{\frametime{n-1}}$ and $\bar{\eventset{}{}}_{\frametime{n}}$ inherently results in the accumulation of drift over time as different regions in the tracked pattern generate events depending on its motion direction in the image plane.
The proposed method includes the use of a refined-on-the-fly template $\bar{\eventset{}{}}_{T}$ that is augmented as new batches are co-registered.
The template is built upon an image-like histogram of the accumulation of all the motion-compensated events that are spatiotemporally close to the track similar to \cite{alzugaray2019,alzugaray2020}.
By binarising and skeletonising it using morphological operations \cite{gonzales2001dip} as shown in Fig.~\ref{fig:dynamic_template}, our method generates virtual events $\hat{\eventset{}{}}_{T}$ that can be used to generate a distance field $\dfield(\coordinate)$.
To robustify the homography estimation, both constraints from $\bar{\eventset{}{}}_{\frametime{n}}$ to $\bar{\eventset{}{}}_{\frametime{n-1}}$ using $\homo{\frametime{n-1}}{\frametime{n}}$, and from $\bar{\eventset{}{}}_{\frametime{n-1}}$ to $\bar{\eventset{}{}}_{\frametime{n}}$ using $\homo{\frametime{n-1}}{\frametime{n}}^{-1}$  are considered in the cost function \eqref{eq:h_cost}.
This prevents the collapse of events into a single location when the pattern texture provides weak constraints on some of the 8 \acp{dof} of the homography.

Starting from a given location $\seed_0$ in the image plane, the position of the track is updated by combining the continuous local \ac{SE2} model and the discrete homographies as $\seed(t) = \trans{t}{\frametime{n\text{-}1}}(t)[\proj{\homo{\frametime{n\text{-}1}}{\frametime{0}}}(\seed_0)] + (t\text{-} \frametime{n\text{-}1})\alpha$ with $\alpha = (\proj{\homo{\frametime{n}}{\frametime{0}}}(\seed_0) -\trans{\frametime{n}}{\frametime{n\text{-}1}}[\proj{\homo{\frametime{n-1}}{\frametime{0}}}(\seed_0)] )/(\frametime{n}\text{-}\frametime{n\text{-}1})$.
The proposed method stops tracking a pattern when the change of marginal likelihood of events after solving \eqref{eq:likelihood_max} is under a certain threshold (proxy for \ac{SE2} motion-compensation failure), when the homography between two successive batches differs significantly from the \ac{SE2} estimate (simple distance threshold between $\proj{\homo{\frametime{n}}{\frametime{0}}}(\seed_0)$ and $\trans{\frametime{n}}{\frametime{n\text{-}1}}[\proj{\homo{\frametime{n-1}}{\frametime{0}}}(\seed_0)]$), or when the pattern is getting close to the image borders.

\section{Experiments}
In this section, we analyse the accuracy of our motion compensation algorithm with simulated data before testing the full tracking pipeline (motion compensation and homography registration) with real-world data.
The same set of parameters is used across all the experiments both in Section \ref{sec:ev_motion} and \ref{sec:tracking}.
The \ac{SE2} \ac{gp} trajectories are modelled with inducing values every 250 events and a square exponential kernel (scaling factor equal to one and a lengthscale three times bigger than the distance between inducing values).
Each batch used for \ac{SE2} motion compensation contains 1250 events.
The lengthscale and scaling factor of the occupancy fields are set to 0.25 and 1 respectively.
The \ac{SE2} local motion compensation is estimated using the BFGS solver and the homography registration is solved with the Levenberg-Marquardt algorithm and a Cauchy loss function.

\subsection{Event-based motion-compensation}
\label{sec:ev_motion}

We provide a quantitative evaluation of the proposed continuous-time \ac{SE2} motion compensation (later referred to as ``c.t. \ac{SE2}'') with a comparison against two other state-of-the-art motion-compensation approaches \cite{stoffregen2019,nunes2021}.
We use simulated data due to the need for accurate groundtruth localisation and scene structure.
The method in \cite{stoffregen2019} relies on an improved version of the popular contrast-maximisation framework \cite{gallego2018contrast} by optimising over a combined objective function that rewards a large accumulation of events at a sparse set of locations.
More closely related to ours, the technique described in \cite{nunes2021} relies on a kernel to quantify a pair-wise ``potential" between events and minimise the dispersion of the motion-compensated events.
Both of these methods have open-source implementations to estimate the 2D translation (referred to as 2-\ac{dof}) in the image plane based on the assumption of constant velocity all along the event batch, but \cite{nunes2021} also provides a \ac{SE2} motion model.
We use the ESIM simulator \cite{rebecq2018} to generate events in two planar environments (a lattice of April Tags and a picture of rocks as textures)  with translation-only, constant-velocity trajectories (suitable for \cite{stoffregen2019} and \cite{nunes2021}-2-\ac{dof}) as well as randomly generated \ac{SE2} trajectories.

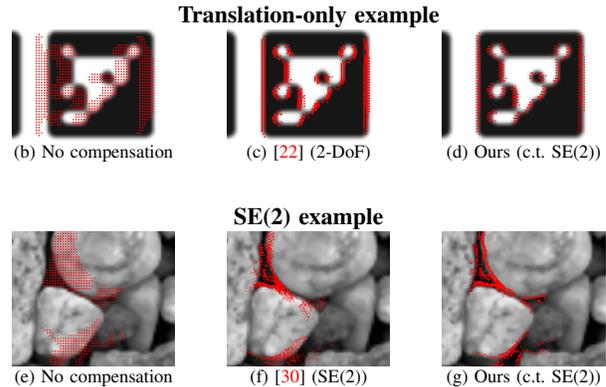
\begin{figure}
    \centering
    \def\imagewidth{0.25\columnwidth}
    \def\widthcol{2.6cm}
    \def\legendspacing{0.05cm}
    \def\vertspacing{1.2cm}
    \def\horispacing{0.7cm}
    \def\titlespacing{0.04cm}
    \def\titleshift{0.0cm}
    \def\colsize{1.9cm}
    \def\colsizeh{1.05cm}
    \newcolumntype{C}[1]{>{\centering\let\newline\\\arraybackslash\hspace{0pt}}m{#1}}
    \begin{scriptsize}
        \begin{tabular}{C{\colsize} || C{\colsizeh} | C{\colsizeh} | C{\colsizeh} | C{\colsizeh}}
            \hline
                        \multirow{2}*{\small Dataset} & \small \cite{stoffregen2019} & \small \cite{nunes2021} & \small \cite{nunes2021}  & Ours \\                        
             & {\scriptsize 2-DoF } & {\scriptsize 2-DoF} & {\scriptsize SE(2)} & {\scriptsize c.t. SE(2)}
            \\
            \hline
            \hline
    April Tags & $1.58\units{px}$ & $2.41\units{px}$& $2.11\units{px}$& $\mathbf{0.25}\units{px}$\\
    2D Translation & ($100\%$) & ($54\%$) & ($58\%$) & ($100\%$)
    \\
    \hline
    April Tags & $4.21\units{px}$ & $2.76\units{px}$ & $2.53\units{px}$ & $\mathbf{0.38}\units{px}$ \\
    SE(2) & ($100\%$) & ($94\%$) & ($98\%$) & ($100\%$)
            \\
            \hline
            \hline
    Rocks & $2.05\units{px}$ & $1.66\units{px}$ & $1.76\units{px}$ & $\mathbf{0.50}\units{px}$ \\
    2D Translation & ($100\%$) & ($78\%$) & ($78\%$) & ($100\%$)\\
    \hline
    Rocks & $6.65\units{px}$ & $3.72\units{px}$ & $2.37\units{px}$ & $\mathbf{0.71}\units{px}$ \\
    SE(2) & ($54\%$) & ($90\%$) & ($90\%$) & ($100\%$)
    \\
    \hline
        \end{tabular}
    \end{scriptsize}
    
    \vspace{0.05cm}
    \scriptsize (a) Motion-compensation accuracy  and success rate in simulated data.
    \vspace{0.7cm}
    
    \begin{tikzpicture}
        \tikzstyle{legend} = [inner sep=0,outer sep=0, rectangle, text width=\widthcol, align=center, anchor=north, font=\scriptsize]
        \tikzstyle{title} = [inner sep=0,outer sep=0, rectangle, text width=0.98\columnwidth, align=center, anchor=south, font=\small]
        \node[inner sep=0,outer sep=0] (aprilraw) {\includegraphics[clip, trim=6cm 12.4cm 6cm 11.1cm, width=\imagewidth]{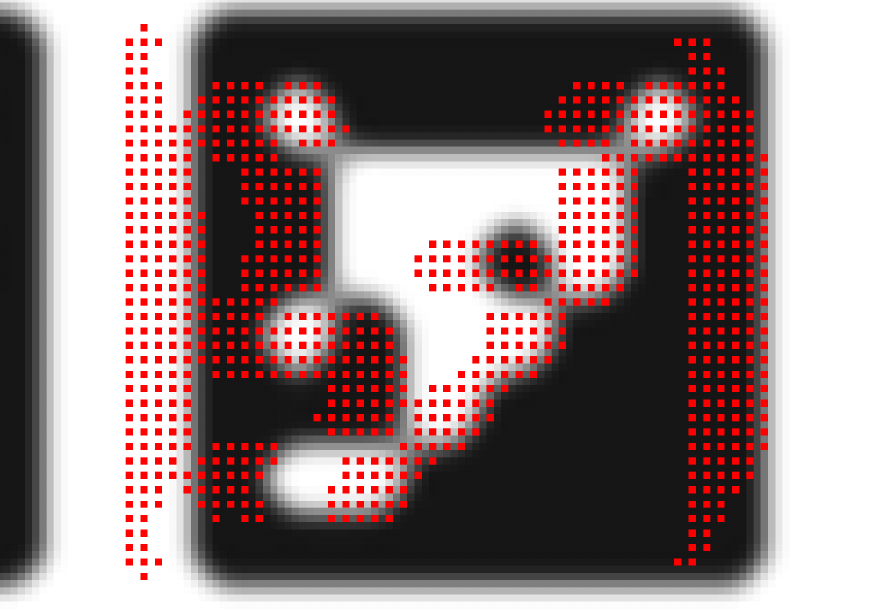}};
        \node[right=\horispacing of aprilraw, inner sep=0,outer sep=0] (aprilcmax) {\includegraphics[clip, trim=6cm 12.4cm 6cm 11.1cm, width=\imagewidth]{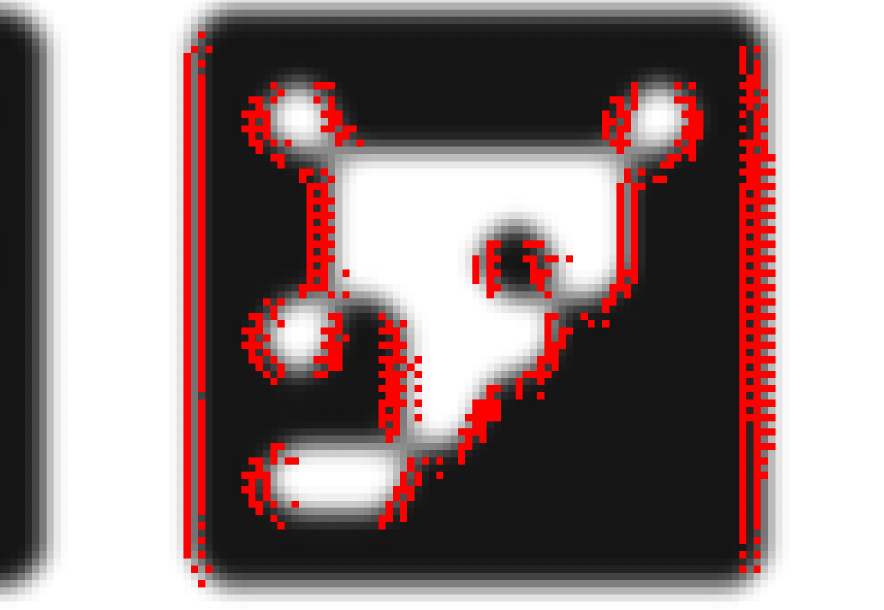}};
        \node[right=\horispacing of aprilcmax, inner sep=0,outer sep=0] (aprilgp) {\includegraphics[clip, trim=6cm 12.4cm 6cm 11.1cm, width=\imagewidth]{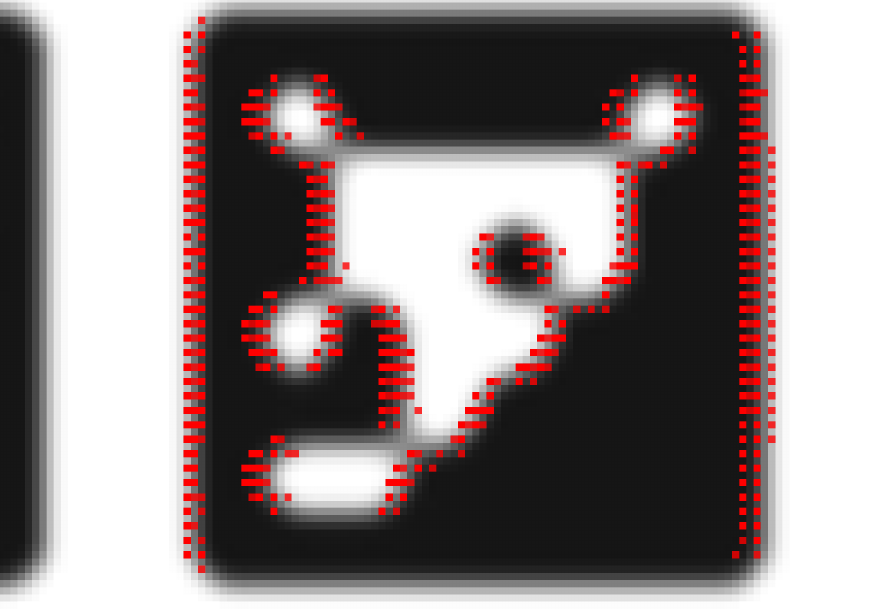}};
        \node[below=\vertspacing of aprilraw, inner sep=0,outer sep=0] (rockraw) {\includegraphics[clip, trim=6cm 11.7cm 6cm 10.6cm, width=\imagewidth]{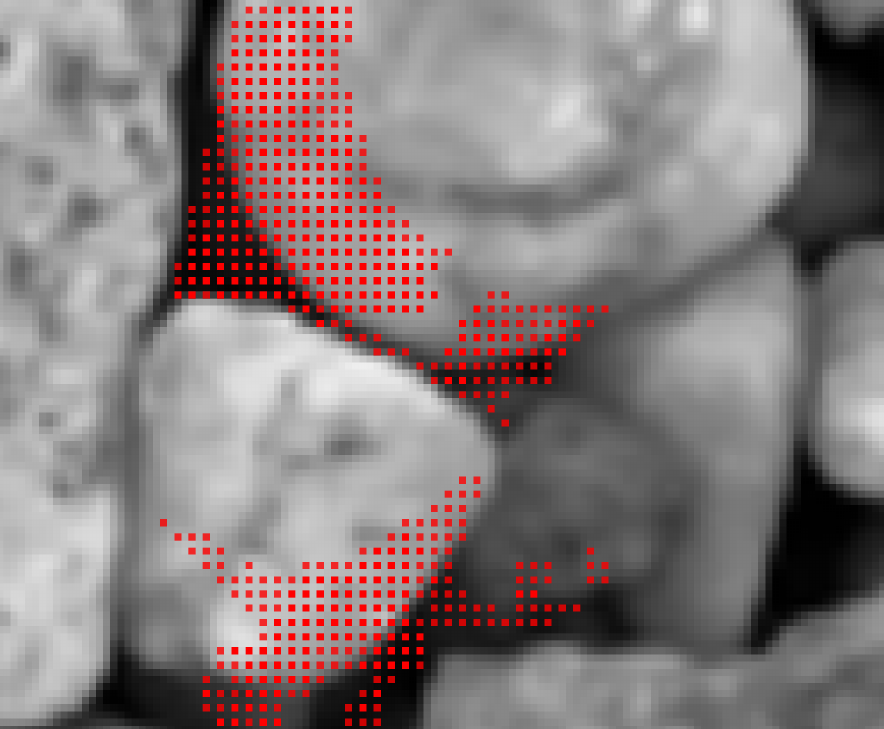}};
        \node[right=\horispacing of rockraw, inner sep=0,outer sep=0] (rockentropy) {\includegraphics[clip, trim=6cm 11.7cm 6cm 10.6cm, width=\imagewidth]{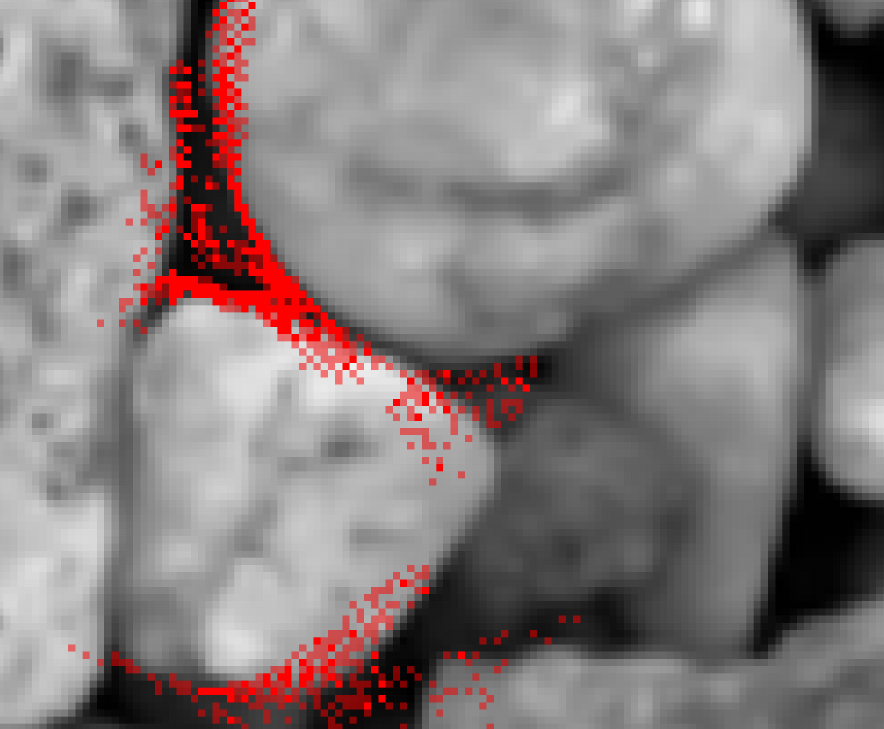}};
        \node[right=\horispacing of rockentropy, inner sep=0,outer sep=0] (rockgp) {\includegraphics[clip, trim=6cm 11.7cm 6cm 10.6cm, width=\imagewidth]{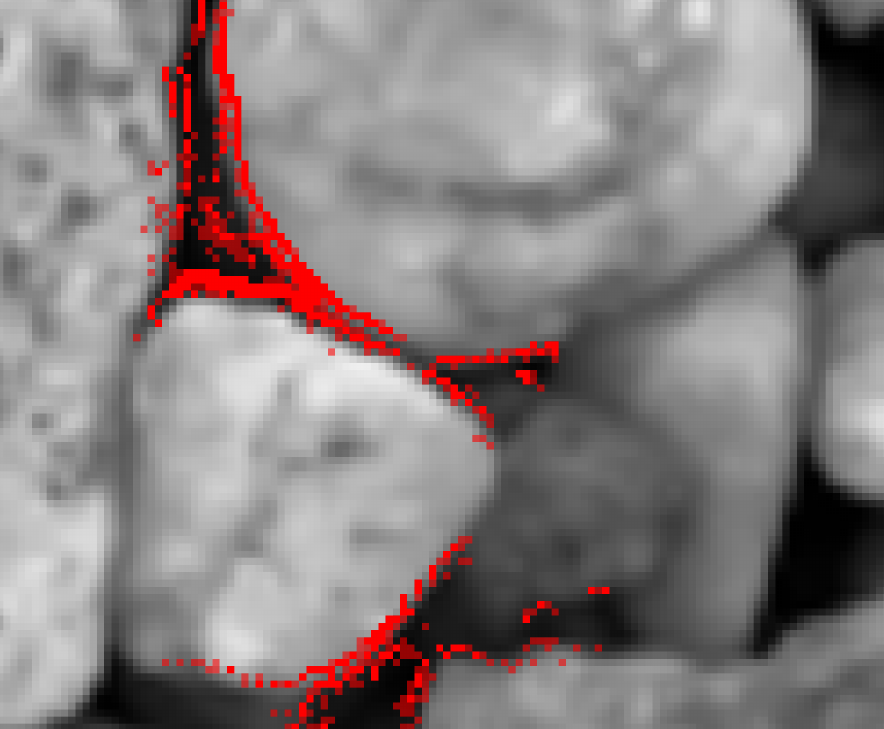}};

        \node[legend, below=\legendspacing of aprilraw]        {(b) No compensation};
        \node[legend, below=\legendspacing of aprilcmax]     {(c) \cite{stoffregen2019} (2-DoF)};
        \node[legend, below=\legendspacing of aprilgp]     {(d) Ours (c.t. SE(2))};
        \node[legend, below=\legendspacing of rockraw]        {(e) No compensation};
        \node[legend, below=\legendspacing of rockentropy]     {(f) \cite{nunes2021} (SE(2))};
        \node[legend, below=\legendspacing of rockgp]     {(g) Ours (c.t. SE(2))};
        \node[title, above=\titlespacing of aprilcmax, xshift=\titleshift]     {\small \textbf{Translation-only example}};
        \node[title, above=\titlespacing of rockentropy, xshift=\titleshift]     {\small \textbf{SE(2) example}};
    \end{tikzpicture}
    \vspace{-0.2cm}
    \caption{Average RMS reprojection error and success rate (RMSE $< 7\units{px}$) of event motion-compensation on simulated data with examples of motion-compensated event batches. (b) and (d) show the raw events (in red) over the corresponding grayscale images. (c,d,f,g) show motion-compensated events. Note that grayscale images are not used in any of the methods.}
    \label{fig:se2_associations}
\end{figure}

To quantify the performance of the different methods, we use the reprojection error between the motion-compensated events and their groundtruth locations based on the known scene structure and camera poses.
The results in Fig.~\ref{fig:se2_associations} have been obtained by averaging the RMSE over the successful runs (out of 50) for each dataset. 
A run is considered successful if the RMSE of all events in the batch is inferior to $7\units{px}$.
When successful, \cite{stoffregen2019} and \cite{nunes2021}-2-\ac{dof} provide similar levels of accuracy but \cite{nunes2021} does not seem to always converge towards a good solution especially in the low-texture environment (lower success rate).
When it comes to \ac{SE2} motion, the accuracy of \cite{stoffregen2019} and \cite{nunes2021}-2-\ac{dof} expectedly decreases whereas \cite{nunes2021}-SE2 retains certain level of accuracy.
The proposed motion-compensation method consistently outperforms the alternatives by an order of magnitude for both trajectory types and across the different scenes.
We believe that the difference in performance between our method and \cite{nunes2021}-\ac{SE2} can be mostly explained by the flexibility of our continuous-time motion model contrasting with the constant velocities assumption in \cite{nunes2021}-\ac{SE2}.

While the proposed approach excels in motion-compensation accuracy compared to the alternatives, our method is computationally more demanding than \cite{stoffregen2019} and \cite{nunes2021} due to the cubic complexity of \eqref{eq:hyperparam_learning}: around $25\units{s}$ per batch with our method while \cite{stoffregen2019} and \cite{nunes2021} yield results in under a second.
A simple way to reduce the complexity of the proposed motion-compensation algorithm is to reduce the number of events used in the optimisation.
As an example, with straightforward uniform downsampling from 1250 to 400 events, our method leads to an error of $0.53\units{px}$ in around $3\units{s}$ for the April Tags \ac{SE2} dataset.
We believe that a more principled approach to downsampling and a multi-scale approach (coarse-to-fine) can significantly improve the computational efficiency of the proposed method.
However, due to space limitation, this paper focuses on demonstrating the soundness of our novel \ac{gp}-based approach to motion compensation and leaves as future work further developments to decrease the computational cost and the analysis of their analysis on the accuracy.


\subsection{Event-based pattern tracking}
\label{sec:tracking}

\begin{figure}[ht]
    \centering
    \includegraphics[width=\columnwidth]{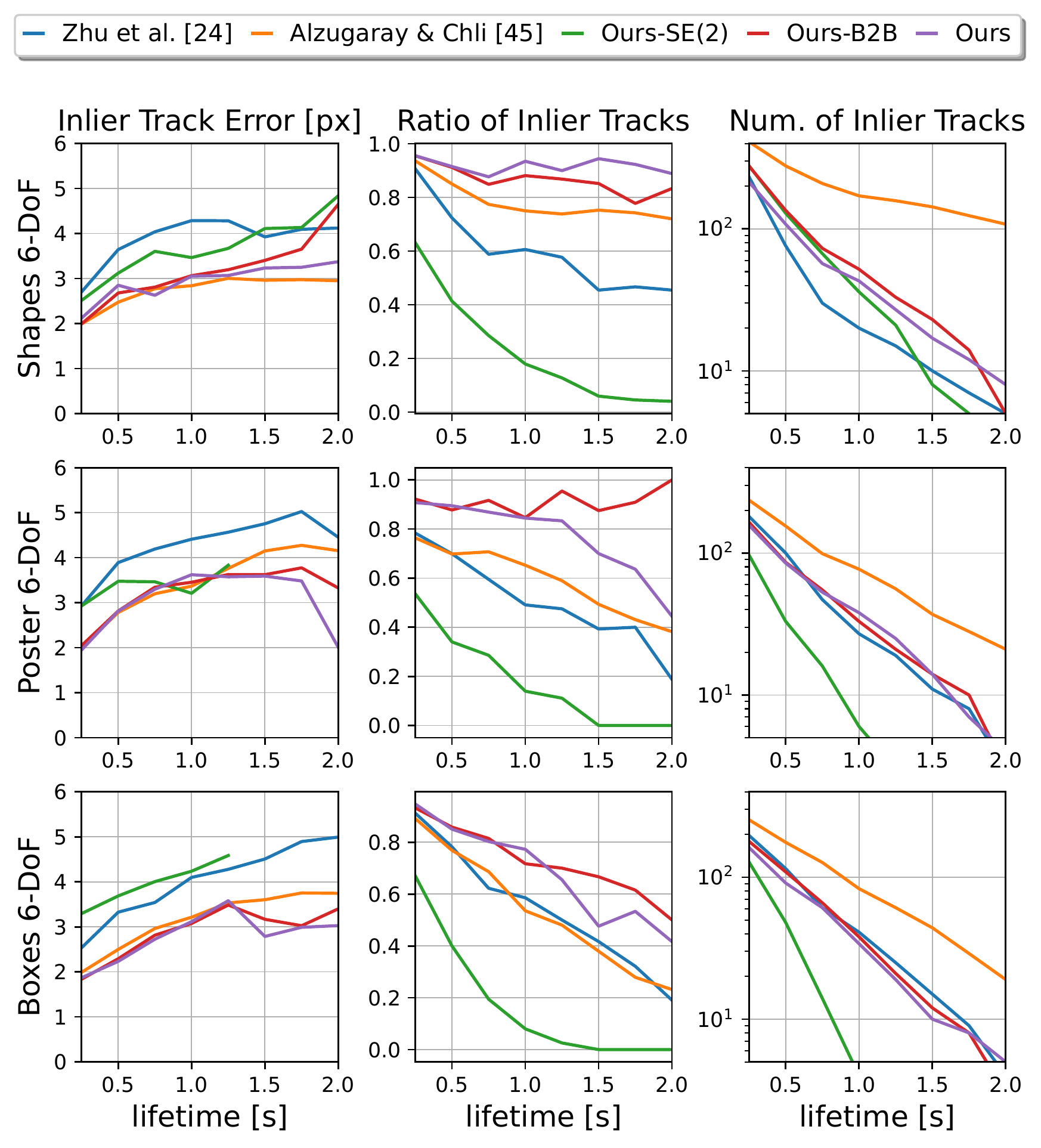}
    \vspace{-0.8cm}
    \caption{Analysis of the proposed tracking pipeline using the KLT-based error metric in different datasets. All the plots are functions of the tracks' lifetime. The first column shows the average RMS error $\downarrow$ (in $\units{px}$) between the inlier tracks and KLT tracks. The second column is the ratio of inlier tracks $\uparrow$ and the third column is the actual number of inlier tracks $\uparrow$.}
    \label{fig:tracking_benchmark}
\end{figure}

We analyse the performance of the full tracking pipeline in real-world scenarios and compare our results with \cite{zhu2017} and \cite{alzugaray2020} on three 6-\ac{dof} datasets from \cite{mueggler2017} (\texttt{shapes\_6dof}, \texttt{poster\_6dof}, and \texttt{boxes\_6dof}).
Each dataset is one minute long and contains increasingly challenging camera motion.
Both \texttt{shapes\_6dof} and \texttt{poster\_6dof} observe planar scenes with simple shape-based and realistic rock-like textures, respectively.
The scene in \texttt{boxes\_6dof} consists of a stack of textured boxes arranged in a pile.
As a real 3D scene, this dataset contains numerous occlusions that can hinder the tracking process.
In the absence of ground truth for pattern tracking evaluation in these datasets, we opt for the KLT-based tracking metric described in \cite{gehrig2020} that compares event-based feature tracks with tracks obtained in frame-based data using the KLT tracker\footnote{Note that, while this metric provides a reasonable ground for comparison across different approaches, the results are completely dependent on the quality of the intensity-based KLT tracks which are easily compromised during motion blur, in poorly illuminated scenarios, or in poorly textured areas. See \cite{alzugaray2018a, alzugaray2019,alzugaray2020,yilmaz2021} for further discussion on the topic.}.
Accordingly, we use only the first 25 seconds of each dataset to limit the KLT tracker performance degradation due to motion blur on intensity images.
As we focus on tracking, we employ the event-based feature detections provided by \cite{zhu2017} to initialise the tracks' starting point of all the compared methods for the sake of fairness.
Atop the comparison with other methods, we conduct an ablation study by considering two additional versions of the proposed tracker that all perform local \ac{SE2} motion compensation but differ in their batch registration strategies between motion-compensated batches of events.
\emph{Ours-\ac{SE2}} uses directly the output of the \ac{SE2} motion compensation to register the events from batch-to-batch (no homography estimation).
\emph{Ours-B2B} corresponds to the proposed pipeline as show in Fig.~\ref{fig:method_diagram} but without the dynamic template (solely batch-to-batch homography).
\emph{Ours} is the full proposed pipeline with the refined-on-the-fly template.

While the proposed method and \cite{zhu2017} are able to detect tracking failures on-the-go, thus, discarding unreliable feature tracks in an online fashion, the method from \cite{alzugaray2020} does not implement such capability.
Accordingly, we create a set of ``inlier tracks" by cutting off tracks when the error goes above $10\units{px}$ similarly to the evaluation in \cite{alzugaray2020}.
Fig.~\ref{fig:tracking_benchmark} reports the RMS tracking error of the inlier tracks (first column) as a function of their lifetime, as well as the ratio of inlier tracks among all the tracks (second column), and the actual number of inlier tracks (third column).
In terms of tracking error, the proposed method and \cite{alzugaray2020} offer similar levels of accuracy across the different scenes, and both significantly outperform \cite{zhu2017}.
\emph{Ours} and \cite{zhu2017} have approximately the same number of inlier tracks. 
While \cite{alzugaray2020} retains the highest number of inlier tracks among all the methods, these inliers are only retrievable in this offline evaluation stage, as \cite{alzugaray2020} does not implement any online outlier rejection policy.
Both \cite{zhu2017} and \cite{alzugaray2020} display a ratio of inliers that rapidly decreases for longer track lifetimes.
Our method keeps this ratio mostly above 80-90\% in \texttt{shapes\_6dof} and \texttt{poster\_6dof}, and above 50\% in \texttt{boxes\_6dof}, indicating a smaller number of outliers in the tracks generated.
This high ratio of inliers is key in downstream applications such as \ac{vo} where only a minimal amount of outliers would have to be handled.
Qualitative results are provided in the associated video.
In summary, compared with existing methods, the proposed framework proved to be more robust (fewer outliers) in most scenarios while performing at a similar level of inlier accuracy.

In the ablation study, \emph{Ours-\ac{SE2}} significantly underperforms the other two versions.
While the inlier tracks' accuracy is similar to those from \cite{zhu2017}, the ratio of inliers rapidly decreases.
The decrease in accuracy of \emph{Ours-\ac{SE2}} compared to \emph{Ours-B2B} and \cite{alzugaray2020} shows the limits of the \ac{SE2} model in the absence of a dynamic template (\cite{alzugaray2020} also estimates a \ac{SE2} motion but leverages a refined-on-the-fly template).
The higher inlier ratio of \emph{Ours-B2B} and \emph{Ours} highlights the performance of our track-termination criteria that confronts the \ac{SE2} and homography estimates.
The use of the proposed refined-on-the-fly template does not seem to significantly affect the performance (\emph{Ours-B2B} vs. \emph{Ours}).
We believe that the drift of our batch-to-batch homography registration is too small to observe any benefits from the refined-on-the-fly template for tracks under $2\units{s}$ of lifetime.

\section{Conclusion}

In this paper, we presented both a novel event-based method for continuous-time motion-compensation built upon \ac{gp} theory, and a tracking algorithm based on homography registration between motion-compensated batches. 
The first component models the occupancy of the events in the image plane with a \ac{gp} that embeds the events' motion in its covariance kernel.
The motion is estimated by maximising the marginal likelihood of the events with respect to the trajectory parameters.
The homography registration leverages \ac{gp} distance fields to minimise distances between motion-compensated events and a template.
Via simulated and real-world experiments, we demonstrated state-of-the-art performance both for motion compensation and pattern tracking.

Future work will address the computational cost of the \ac{SE2} motion-compensation (around 95\% of the tracking computation time) by including the use of structured-kernel interpolation \cite{wilson2015kissgp} which can significantly reduce the computational complexity with negligible impact on accuracy as demonstrated in \cite{giubilato2020towards}.




\bibliographystyle{IEEEtran}
\bibliography{bibliography, bibliography_events}

\begin{acronym}[AAAAAAAAA]
    \acro{1d}[1D]{One-Dimensional}
    \acro{2d}[2D]{Two-Dimensional}
    \acro{3d}[3D]{Three-Dimensional}
    \acro{cas}[CAS]{Centre for Autonomous Systems}
    \acro{cpu}[CPU]{Central Processing Unit}
    \acro{dof}[DoF]{Degree-of-Freedom}
    \acro{dvs}[DVS]{Dynamic Vision Sensor}
    \acrodefplural{dvs}[DVS's]{Dynamic Vision Sensors}
    \acro{ekf}[EKF]{Extended Kalman filter}
    \acro{fov}[FoV]{Field-of-View}
    \acro{gnss}[GNSS]{Global Navigation Satellite System}
    \acrodefplural{gnss}[GNSS's]{Global Navigation Satellite Systems}
    \acro{gp}[GP]{Gaussian Process}
    \acrodefplural{gp}[GPs]{Gaussian Processes}
    \acro{gpm}[GPM]{Gaussian Preintegrated Measurement}
    \acro{ugpm}[UGPM]{Unified Gaussian Preintegrated Measurement}
    \acro{gps}[GPS]{Global Position System}
    \acrodefplural{gps}[GPS's]{Global Position Systems}
    \acro{gpu}[GPU]{Graphic Processing Unit}
    \acro{hdr}[HDR]{High Dynamic Range}
    \acro{icp}[ICP]{Iterative Closest Point}
    \acro{idol}[IDOL]{IMU-DVS Odometry using Lines}
    \acro{imu}[IMU]{Inertial Measurement Unit}
    \acro{in2laama}[IN2LAAMA]{INertial Lidar Localisation Autocalibration And MApping}
    \acro{kf}[KF]{Kalman Filter}
    \acro{kl}[KL]{Kullback¿Leibler}
    \acro{lidar}[LiDAR]{Light Detection And Ranging Sensor}
    \acro{lpm}[LPM]{Linear Preintegrated Measurement}
    \acro{map}[MAP]{Maximum A Posteriori}
    \acro{mle}[MLE]{Maximum Likelihood Estimation}
    \acro{ndt}[NDT]{Normal Distribution Transform}
    \acro{pm}[PM]{Preintegrated Measurement}
    \acro{rrbt}[RRBT]{Rapidly Exploring Random Belief Trees}
    \acro{rgb}[RGB]{Red-Green-Blue}
    \acro{rgbd}[RGBD]{Red-Green-Blue-Depth}
    \acro{rms}[RMS]{Root Mean Squared}
    \acro{rmse}[RMSE]{Root Mean Squared Error}
    \acro{sde}[SDE]{Stochastic Differential Equation}
    \acro{SE3}[SE(3)]{Special Euclidean group in three dimensions}
    \acro{SE2}[SE(2)]{Special Euclidean group in 2D}
    \acro{slam}[SLAM]{Simultaneous Localisation And Mapping}
    \acro{so3}[$\mathfrak{so}$(3)]{Lie algebra of special orthonormal group in three dimensions}
    \acro{SO3}[SO(3)]{Special Orthonormal rotation group in three dimensions}
    \acro{upm}[UPM]{Upsampled-Preintegrated-Measurement}
    \acro{uts}[UTS]{University of Technology, Sydney}
    \acro{vi}[VI]{Visual-Inertial}
    \acro{vio}[VIO]{Visual-Inertial Odometry}
    \acro{vo}[VO]{Visual Odometry}
    \acro{klt}[KLT]{Kanade–Lucas–Tomasi}	
	\acro{em}[EM]{Expectation-Maximization}
 
\end{acronym}

\end{document}